\documentclass[11pt,a4paper]{article}
\usepackage[margin=1in]{geometry}
\usepackage[utf8]{inputenc}
\usepackage[T1]{fontenc}
\usepackage{amsmath,amssymb,amsthm}
\usepackage{enumitem}
\usepackage{hyperref}
\usepackage{longtable}
\usepackage{booktabs}
\usepackage{xcolor}
\usepackage{graphicx}

\hypersetup{colorlinks=true,linkcolor=blue,urlcolor=blue,citecolor=blue}

\title{Beyond Sentiment:\\
Comparing Traditional NLP and LLM-Based\\
Multi-Dimensional Analysis for Political News Evaluation}

\author{
Maryam Fooladi\textsuperscript{1},
Federico Bottino\textsuperscript{1}\\[4pt]
\textsuperscript{1}Kakashi Ventures Accelerator (KVA) / Newjee
}
\date{}

\begin{document}
\maketitle

\begin{abstract}
Traditional sentiment analysis (SA) models, while effective for polarity classification, provide
limited insight into the rhetorical, ideological, and framing dimensions of political
discourse---dimensions that are central to research in the social sciences and humanities (SSH).
In this paper, we present a comparative study of RoBERTa-based sentiment analysis and an
LLM-based multi-dimensional framing analysis platform applied to a corpus of 50 political news
articles from 17 international media outlets. The results reveal a critical limitation we term
\emph{neutral collapse}: RoBERTa classifies 70\% of articles as neutral, effectively flattening
substantively rich political content into an analytically uninformative category. We find that
23\% of neutral-classified articles exhibit negative probability scores above 0.30. By contrast,
the LLM-based approach captures political bias direction and intensity, sensationalism, emotional
appeal, and political framing---yielding multi-dimensional analytical outputs aligned with SSH
epistemologies. We argue that for political media analysis, traditional SA alone is insufficient,
and that LLM-based multi-dimensional frameworks offer a more epistemologically adequate
computational lens for SSH research needs.
\end{abstract}

\noindent\textbf{Keywords:} sentiment analysis, political framing, LLM, media analysis, social
sciences and humanities, multi-dimensional text analysis

\section{Introduction}

The computational analysis of political discourse has become increasingly central to research
across the social sciences and humanities (SSH). Scholars in political science, media studies,
communication, and digital humanities routinely seek to understand not merely \emph{what} is said
in political coverage, but \emph{how} it is said---through which rhetorical frames, with what
ideological orientation, and at what level of emotional intensity~\cite{entman1993framing,
scheufele1999framing}. These epistemological concerns---rooted in framing theory, agenda-setting
research, and critical discourse analysis---define how SSH scholars conceptualise media influence
on public opinion and political behaviour.

Sentiment analysis (SA) has been widely adopted as a first-order computational tool for media
analysis. Transformer-based models such as BERT~\cite{devlin2019bert} and
RoBERTa~\cite{liu2019roberta} achieve state-of-the-art performance on sentiment classification
benchmarks, and specialised variants like the Cardiff NLP Twitter-RoBERTa
model~\cite{loureiro2022timelms} are commonly used in computational social science. However, these
models fundamentally reduce textual content to a three-class polarity judgment---positive, neutral,
or negative---a reduction that may be systematically misaligned with how SSH researchers
conceptualise and analyse political discourse.

The emergence of Large Language Models (LLMs) offers a potential bridge between computational text
analysis and SSH research needs. LLMs can be prompted to perform multi-dimensional analysis that
captures framing strategies, ideological bias, sensationalism, and narrative patterns---categories
that directly correspond to established SSH analytical frameworks~\cite{gilardi2023chatgpt,
ziems2024llmscss, tornberg2024llms}. This raises a critical question for the LLMs4SSH community:
to what extent does traditional sentiment analysis capture the dimensions of political text that
matter to SSH researchers, and can LLM-based approaches better serve these research needs?

In this paper, we address this question through a direct comparative study. We apply both
RoBERTa-based sentiment analysis and an LLM-based multi-dimensional analysis platform to the same
corpus of 50 political news articles from 17 international outlets. The contributions are
threefold:

\begin{enumerate}
  \item We identify and characterise a phenomenon we call \emph{neutral collapse}---the
  systematic tendency of SA models to classify substantively rich political content as
  ``neutral.''
  \item We demonstrate through case-level analysis how this collapse discards information that is
  analytically central to SSH research.
  \item We provide a structured comparison of what each approach offers, arguing for
  complementary rather than substitutive use of these tools.
\end{enumerate}

\section{Related Work}

\subsection{Sentiment Analysis in Political Text}

Sentiment analysis has a long trajectory in NLP, evolving from lexicon-based
approaches~\cite{hu2004mining} through machine learning methods to the current transformer-based
state of the art~\cite{liu2019roberta}. In political text analysis, SA has been applied to social
media discourse around elections~\cite{barbieri2020tweeteval, budiharto2018prediction},
parliamentary debate transcripts~\cite{abercrombie2020sentiment}, and news media
coverage~\cite{hamborg2019automated}. The Cardiff NLP Twitter-RoBERTa
model~\cite{loureiro2022timelms}, fine-tuned on approximately 124 million tweets, has become one
of the most widely cited SA models in computational social science.

However, the suitability of sentiment polarity as a proxy for political text analysis has been
questioned. Van Atteveldt et al.~\cite{vanatteveldt2021validity} conducted a systematic comparison
of SA methods in communication research and found that automated approaches often fail to capture
the relevant variation in political news. Baden et al.~\cite{baden2022gaps} argued that the
conceptual gap between what NLP tools measure and what communication researchers need represents
a fundamental methodological challenge for computational social science.

\subsection{Framing Theory and Multi-Dimensional Analysis}

In political communication, the concept of framing---the selection and emphasis of particular
aspects of perceived reality to promote a particular interpretation~\cite{entman1993framing}---has
been a dominant theoretical framework for over three decades. Scheufele~\cite{scheufele1999framing}
distinguished between media frames and individual frames, while subsequent work identified specific
framing devices including conflict framing, human interest framing, economic consequences framing,
and morality framing~\cite{semetko2000framing}. These multi-dimensional analytical categories bear
little resemblance to the positive-neutral-negative trichotomy of sentiment analysis.

Computational approaches to framing detection have emerged in NLP~\cite{card2015media,
field2018framing}, but typically require pre-defined frame taxonomies and annotated training data.
The rigid category schemes of supervised framing models can miss emergent or context-specific
frames that qualitative SSH researchers would readily identify.

\subsection{LLMs for SSH Research}

Recent work has demonstrated the potential of LLMs for nuanced text analysis tasks relevant to
SSH. Gilardi et al.~\cite{gilardi2023chatgpt} showed that ChatGPT can match or outperform
crowd-annotated data for text classification tasks, including political stance detection.
Ziems et al.~\cite{ziems2024llmscss} provided a comprehensive survey of LLM applications in
computational social science, noting their capacity for flexible, multi-dimensional text
interpretation. T\"ornberg~\cite{tornberg2024llms} argued that LLMs represent a paradigm shift
for social science text analysis by enabling interpretive analysis---going beyond pre-defined
categories to provide contextualised readings of text.

Crucially, Bail~\cite{bail2024generativeai} has argued that the social sciences need AI tools
configured for social-scientific reasoning, not merely repurposed NLP benchmarks. This study
contributes to the discussion by providing direct empirical evidence of the gap between SA and
SSH-relevant analysis, and by demonstrating how LLM-based platforms can bridge it.

\section{Methodology}

\subsection{Corpus Construction}

We constructed a corpus of 50 English-language political news articles published between June and
July 2025, collected from 17 international media outlets. The outlets were selected to represent
diversity along three axes:

\begin{enumerate}
  \item \textbf{Geographic scope} --- including Western Anglophone (BBC, The Guardian, NPR,
  NYTimes, Fox News, NBC News, CNBC), European (France 24, Euronews, Politico.eu, Reuters), and
  Global South/Other outlets (Al Jazeera, Iran International, VOA, CAN, WION).
  \item \textbf{Editorial orientation} --- ranging from outlets commonly associated with
  centre-left editorial positions to those associated with centre-right or explicitly conservative
  stances.
  \item \textbf{Thematic coverage} --- spanning parliamentary governance, diplomatic negotiations,
  election law, civil rights, geopolitical tensions, immigration policy, and corruption.
\end{enumerate}

This corpus design reflects the type of heterogeneous, multi-source political coverage that SSH
researchers routinely encounter and analyse. The intentional diversity of outlets and topics makes
it a challenging test case for any computational analysis tool.

\subsection{RoBERTa Sentiment Analysis}

For the traditional SA approach, we employed the \texttt{cardiffnlp/twitter-roberta-base-sentiment-latest}
model~\cite{loureiro2022timelms}, a RoBERTa-base architecture (125M parameters) fine-tuned on
approximately 124 million tweets for three-class sentiment classification. This model was selected
as it represents the most widely cited open-source SA model in current NLP research.

For each article, the full text was tokenised and truncated to 512 tokens (the model's maximum
input length). The model produces three probability scores via softmax---$P(\text{negative})$,
$P(\text{neutral})$, $P(\text{positive})$---with the highest-scoring class assigned as the
dominant label. We additionally compute a compound score $(P(\text{positive}) - P(\text{negative}))$
and the classification margin (difference between the top two probability scores) to characterise
the model's confidence.

\subsection{LLM-Based Multi-Dimensional Analysis}

For the multi-dimensional analysis, we employed an LLM-based platform that implements a structured
analytical pipeline. Unlike RoBERTa's single-dimension polarity output, this platform processes
full article texts (without truncation) and analyses each article across multiple dimensions
corresponding to established SSH analytical categories.

\paragraph{Political Bias Assessment.}
Each article receives a directional label (left, neutral, right) accompanied by a continuous
intensity score (0--100). This directly addresses a core SSH research question---ideological
orientation of coverage---using a representation that supports both categorical and continuous
analysis.

\paragraph{Sensationalism.}
A continuous score (0--100) measuring the degree of emotional exaggeration, clickbait patterns,
hyperbolic language, and excessive rhetorical appeals. This dimension corresponds to longstanding
concerns in journalism studies about media quality~\cite{molekkozakowska2013sensationalism}.

\paragraph{Emotional Appeal.}
A continuous score (0--100) quantifying the extent to which the article employs emotional rather
than rational argumentation strategies.

\paragraph{Political Framing.}
A composite score (0--100) indicating the overall intensity of political framing devices, including
fear appeal, scapegoating, us-vs-them dichotomies, and victim/hero narratives---categories drawn
from framing theory~\cite{entman1993framing} and propaganda
analysis~\cite{jowett2019propaganda}. The system processes the complete article text, preserving
the full argumentative structure that is lost when text is truncated to 512 tokens.

\section{Results}

\subsection{RoBERTa Sentiment Distribution}

The most striking finding is the overwhelming dominance of the neutral label. Of the 50 political
articles, 35 (70\%) were classified as neutral, 13 (26\%) as negative, and only 2 (4\%) as
positive. The mean probability scores across the corpus were $P(\text{neg}) = 0.291$ ($\sigma = 0.244$),
$P(\text{neu}) = 0.638$ ($\sigma = 0.217$), and $P(\text{pos}) = 0.070$ ($\sigma = 0.124$). The
positive class is effectively suppressed across the entire corpus. Figure~\ref{fig:roberta-dist}
illustrates this distribution.

\begin{figure}[h]
\centering
\fbox{\parbox{0.85\textwidth}{\centering\vspace{1em}
\textit{[Figure placeholder: Panel A --- bar chart of label distribution (Negative 13, Neutral 35,
Positive 2). Panel B --- boxplots of P(neg), P(neu), P(pos) score distributions.]}
\vspace{1em}}}
\caption{RoBERTa sentiment classification results. Panel~A: Label distribution showing neutral
dominance (70\%). Panel~B: Score distributions across the three classes.}
\label{fig:roberta-dist}
\end{figure}

\subsection{The Neutral Collapse Phenomenon}

We term this distributional pattern \emph{neutral collapse}: the systematic tendency of sentiment
models to assign a neutral label to political news articles that are, from an SSH perspective,
substantively rich in framing, bias, and rhetorical strategy. Neutral collapse occurs because
journalistic writing employs hedging, attribution, balanced sourcing, and formal
register---linguistic features that sentiment models interpret as absence of valence, even when the
underlying content carries significant ideological and rhetorical weight.

The problem is quantitatively characterised by two findings. First, 8 of the 35 neutral-classified
articles (23\%) have negative probability scores above 0.30, indicating that the model detects
substantial negative content but is narrowly overruled by the neutral probability. Second,
classification confidence is often low: 3 articles (6\%) have a margin of less than 0.10 between
the top two classes, 6 (12\%) have margins below 0.15, and 8 (16\%) have margins below 0.20.

Figure~\ref{fig:cases} presents three illustrative cases. An article on the ouster of Australia's
first female Liberal Party leader (BBC) received a neutral label despite a negative score of 0.48
against a neutral of 0.50---a margin of 0.02. An SSH researcher analysing gender and political
leadership would find this ``neutral'' classification misleading. Similarly, articles on
Mexican-UK diplomatic tensions ($P(\text{neg}) = 0.43$) and Franco-German military disagreements
($P(\text{neg}) = 0.41$) were classified as neutral despite containing substantial political
conflict.

\begin{figure}[h]
\centering
\fbox{\parbox{0.85\textwidth}{\centering\vspace{1em}
\textit{[Figure placeholder: three horizontal bar charts showing per-class probabilities for
the BBC, Guardian, and Politico.eu cases, with margin annotations and the common ``All classified
$\to$ Neutral'' header.]}
\vspace{1em}}}
\caption{Case studies of neutral-classified articles with high negative scores, illustrating
neutral collapse at decision boundaries.}
\label{fig:cases}
\end{figure}

\subsection{Cross-Outlet Patterns}

When sentiment scores are aggregated by outlet region, RoBERTa reveals limited but notable patterns
in the negative dimension. European outlets (mean $P(\text{neg}) = 0.374$) scored highest, followed
by Western Anglophone outlets (0.293) and Global South/Other outlets (0.258). However, the positive
dimension shows almost no inter-outlet variation, and the neutral label dominates uniformly across
all regions (62--74\%). Figure~\ref{fig:outlets} shows the per-outlet breakdown.

\begin{figure}[h]
\centering
\fbox{\parbox{0.85\textwidth}{\centering\vspace{1em}
\textit{[Figure placeholder: horizontal stacked bar chart of mean P(neg)/P(neu)/P(pos) per outlet,
ordered by P(neg) score.]}
\vspace{1em}}}
\caption{Stacked average sentiment scores by outlet, ordered by negative score. The dominance of
neutral (grey) is uniform across outlets.}
\label{fig:outlets}
\end{figure}

This means that RoBERTa provides a researcher with essentially one useful dimension of
variation---negative intensity---while the neutral and positive categories offer minimal analytical
discriminability for political text.

\subsection{LLM-Based Multi-Dimensional Results}

In contrast to RoBERTa's single-axis output, the LLM-based platform provides analytically
differentiated outputs across four continuous dimensions plus a categorical bias direction. The
\textit{political bias assessment} distributes articles across left, neutral, and right categories
with accompanying intensity scores, directly addressing the question: from which political
perspective is content presented? Unlike sentiment polarity, which conflates ideological direction
with emotional valence, the bias dimension preserves directional information central to SSH
analysis.

The \textit{sensationalism} and \textit{emotional appeal} dimensions capture rhetorical strategies
that operate orthogonally to sentiment polarity. An article may be classified as ``neutral'' by
RoBERTa while simultaneously exhibiting high sensationalism, because journalistic sensationalism is
achieved through narrative structure and topic selection rather than through lexical polarity
cues.

The \textit{political framing} dimension identifies rhetorical strategies---fear appeal,
scapegoating, us-vs-them framing, victim/hero narratives---that constitute the core object of study
in framing theory~\cite{entman1993framing, semetko2000framing}. These devices lie entirely outside
the scope of polarity-based SA.

\subsection{Dimensional Comparison}

The comparison reveals a fundamental asymmetry: RoBERTa provides one dimension (polarity) with high
reproducibility but low SSH relevance, while the LLM-based approach provides four continuous
dimensions plus categorical bias direction, all with direct correspondence to established SSH
analytical frameworks. A particularly important structural difference is input length: RoBERTa's
512-token truncation means that for a typical article of 800--1500 words, the model analyses only
the first third to half. Political framing often operates through cumulative article structure,
which is preserved only by full-text processing.

\section{Discussion}

\subsection{Implications for SSH Research}

The findings point to a fundamental misalignment between what traditional SA measures and what SSH
researchers need. The neutral collapse phenomenon is not a failure of RoBERTa---the model performs
as designed, classifying emotional polarity---but rather reflects the fact that political news
articles are not primarily characterised by lexical emotional polarity. They are characterised by
framing, ideological orientation, and rhetorical strategy---precisely the dimensions the LLM-based
approach captures.

This has concrete implications for research design. The widespread use of SA as a proxy for media
analysis~\cite{vanatteveldt2021validity} risks systematic information loss. When 70\% of a corpus
is classified as ``neutral,'' the researcher is left with a category that provides no analytical
leverage. SSH researchers who rely on SA as their primary tool may be drawing conclusions from only
the most extreme 30\% of their corpus, missing the nuanced framing that characterises the majority
of political coverage.

\subsection{Complementary Rather Than Substitutive Use}

We do not argue that LLM-based analysis should replace SA. Each approach has distinctive strengths.
RoBERTa offers high reproducibility (being open-source with deterministic outputs), low
computational cost, and a well-understood metric useful for large-scale longitudinal studies. For
research questions concerning emotional valence---such as tracking public mood---SA remains
appropriate.

However, for research questions concerning framing, ideology, or media quality---the core concerns
of political communication scholarship---SA alone is insufficient. We propose a complementary
framework: SA provides a first-pass polarity screen, and LLM-based multi-dimensional analysis
provides the deeper, SSH-aligned analytical layer.

\subsection{Challenges and Considerations}

The LLM-based approach introduces its own challenges. First, \emph{reproducibility}: proprietary
models may produce slightly different outputs across API versions. Second, \emph{cost}: LLM
inference is orders of magnitude more expensive than running a fine-tuned transformer. Third,
\emph{validation}: while the LLM-based outputs are intuitively richer, their alignment with expert
human judgments requires systematic evaluation.

A further consideration concerns the \emph{opacity of LLM reasoning}. LLM-based analysis produces
more interpretable \emph{outputs} (bias labels, framing categories) but relies on more opaque
internal \emph{processes}. For SSH researchers accustomed to transparent methodology, this
trade-off deserves careful consideration.

\subsection{Toward SSH-Aligned NLP Tools}

These findings contribute to a broader argument~\cite{bail2024generativeai, tornberg2024llms}: the
social sciences need AI tools configured for social-scientific reasoning, not merely repurposed
from NLP benchmarks. The LLMs4SSH community is well positioned to develop shared evaluation
frameworks assessing tools not by NLP benchmark accuracy but by SSH research utility. We suggest
criteria including: (1) dimensional coverage of SSH-relevant categories, (2) sensitivity to framing
and rhetorical features, (3) robustness across journalistic registers, and (4) alignment with
expert SSH judgments.

\section{Conclusion}

We have presented a comparative study demonstrating that RoBERTa-based sentiment analysis suffers
from neutral collapse when applied to political news articles---classifying 70\% of substantively
rich content as neutral, with 23\% of neutral-classified articles exhibiting borderline negative
scores above 0.30. An LLM-based multi-dimensional platform captures bias direction and intensity,
sensationalism, emotional appeal, and political framing, providing analytically richer outputs
aligned with SSH research needs.

The findings carry three practical implications. First, researchers should be cautious about using
SA as the sole tool for political media analysis. Second, LLM-based multi-dimensional analysis
offers a promising complement to SA, provided validation against expert SSH judgments is conducted.
Third, the development of SSH-aligned evaluation frameworks---measuring tool utility by research
relevance rather than NLP benchmark performance---is an important direction for the community.

Future work will extend this comparison to larger, multilingual corpora; incorporate expert
evaluation by political communication scholars; investigate whether fine-tuning open-source LLMs
can approximate the multi-dimensional analysis currently achieved with proprietary models; and
explore how the complementary use of SA and LLM-based analysis can be operationalised in SSH
research workflows.

\section*{Ethics Statement}

All articles in the corpus were drawn from publicly available news sources. No personally
identifiable information was collected. The use of an LLM-based analysis platform raises ethical
considerations: large model inference has non-trivial environmental cost, and reliance on a
proprietary model introduces dependencies on commercial infrastructure.

\section*{Limitations}

The corpus of 50 articles, while diverse, is small and limits statistical robustness. The LLM
platform's reliance on a proprietary model means exact replication depends on API availability. The
absence of systematic human expert evaluation is a significant gap. RoBERTa's 512-token truncation
places it at a structural disadvantage, though this constraint is shared by most current
transformer-based SA models. Finally, the analysis focuses on English-language political news;
generalisability to other languages remains to be established.


\end{document}